\documentclass[letterpaper, 10 pt, conference]{ieeeconf}  

\IEEEoverridecommandlockouts                              

\usepackage{soul}
\usepackage{graphics}
\usepackage{graphicx}
\usepackage{color}
\usepackage{xcolor}
\usepackage{booktabs}
\usepackage{algorithm}
\usepackage{algorithmic}
\usepackage{CJK}
\usepackage{amsmath}
\usepackage{geometry}   
\title{\LARGE \bf
PointIT: A Fast Tracking Framework Based on 3D Instance Segmentation}

\author{Yuan Wang \ \ \ \
        Yang Yu \ \ \ \
        Ming Liu
\thanks{The Hong Kong University of Science and Technology,
        {\tt\small \{ywangeq, yang.yu, eelium\}connect.ust.hk}}%
}

\begin{document}

\maketitle
\thispagestyle{empty}
\pagestyle{empty}

\begin{abstract}
Recently most popular tracking frameworks focus on 2D image sequences. They seldom track the 3D object in point clouds.
In this paper, we propose \textit{PointIT}, a fast, simple tracking method based on 3D on-road instance segmentation. 
Firstly, we transform 3D LiDAR data into the spherical image with the size of $\boldsymbol{64 \times512 \times4}$  and feed it into instance segment model to get the predicted instance mask for each class. 
Then we use \textit{MobileNet} as our primary encoder instead of the original \textit{ResNet} to reduce the computational complexity. 
Finally, we extend the Sort algorithm with this instance framework to realize tracking in the 3D LiDAR point cloud data. 
The model is trained on the spherical images dataset with the corresponding instance label masks which are provided by KITTI 3D Object Track dataset. 
According to the experiment results, our network can achieve on Average Precision (AP) of 0.617 and the performance of multi-tracking task has also been improved.

\end{abstract}

\section{INTRODUCTION}
\subsection{Motivation}
With the recent development of deep learning, many object detection, classification and semantic segmentation methods have achieved impressive performance in many areas, such as autonomous vehicles (AV) and advanced driver-assistance systems (ADAS). 
The object track frameworks play an essential role in both systems. In order to realize object tracking, many popular algorithms have been proposed based on the object detection frameworks, such as Sort \cite{sort}, Deepsort \cite{deepsort} and ASMS \cite{vojir2014robust} which all achieved the excellent performance.
However, these tracking frameworks perform in the 2D image and track the target based on the stable 2D object detectors. 
Most of them take the Intersection-over-Union (IoU) matching matrix or other data association methods to match the detected objects in the continuous frames, which means that these trackers are often influenced by the performance of the detectors and not robust at solving the interaction problem. 

The most fundamental aspect of intelligent vehicle applications is to help the vehicles understand the surrounding environment and make a firm decision based on that information. 
We think the 3D point cloud is more accurate and stable than the image in the perception of environment. The point clouds can be gathered easily by LiDAR which is usually equipped on intelligent vehicles. 
Also, the scale of an object in point clouds is invariant which is always changed in the common RGB image. 
There are many popular 3D object detection models, such as Complex-Yolo \cite{Complex-YOLO}, MV3D \cite{MV3D}, VoxelNet \cite{VoxelNet} and PIXOR \cite{PIXOR} available if we want to transform the similar idea of the IoU matching matric from the 2D bounding box to the 3D bounding box. Although computing the IoU matching matrix based on 3D bounding box can work, they still have the challenges that those 3D object detection models do not perform well in the prediction of the orientation of the object, 
which means the IoU matrix will be built with a high error based on the 3D object proposal, even those models can locate the 3D object in the space precisely.
\subsection{Contribution}
To solve this problem, we firstly project 3D point clouds acquired using LiDAR into the spherical coordinate system inspired by SqueezeSeg \cite{SqueezeSeg} and PointSeg \cite{PointSeg}. 
By this 2D data, we can perform 2D tracking on it and recover the 3D information directly. 
To mitigate the interaction problem in common track frameworks, we introduce the additional 3D information from the 2D projected spherical image by the 3D instance segmentation framework. Compared with the 3D object detection, the 3D instance segmentation is more suitable to locate the object in the 3D space, because it not only locates the object, but also predicts the point-wise mask for the further process. 
In spite of this, a trade-off between the cost time in the 3D data feature extraction and prediction accuracy always exists in 3D instance segmentation task. 
Owning to that many 3D instance segmentation tasks use the Multi-layer Perceptron (MLP) to generate the feature representations which cannot be applied well in a large scale scene directly and have the problem on time-consuming due to a large number of concatenations of PointNet \cite{pointnet} layers.

We propose an efficient 3D instance segmentation framework with the light weighted network structure which takes the advantages of the spherical image to speed up the processing,
by compressing the 3D information in 2D data type with channels.
We extend the fast-tracking algorithm Sort \cite{sort} to take 3D information into consideration when builds the cost matrix and achieves the 3D tracking framework for the road objects. 
We name this pipeline as \textit{PointIT}, as shown in Fig. \ref{fig:pipeline}.
Our work can track 3D objects at a speed of 15 fps in the entire forward process. In general, we highlight our contributions as follows:

\begin{figure*}[!ht]
  \centering
    \includegraphics[width=1\textwidth]{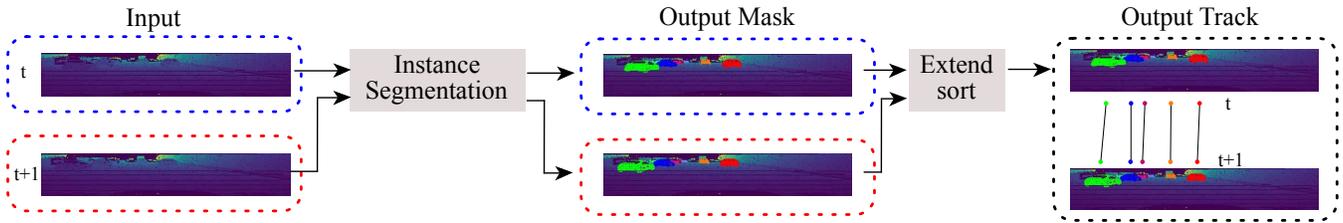}
  \caption{The pipeline of \textit{PointIT}. It contains two part: (1) instance segmentation process which gets the instance segmentation from the input of a spherical image; (2) generation the association matrix with extended sort from the output mask and corresponding space information. The different colors in this figure are only set to separate all instances. 
}
  \label{fig:pipeline}
\end{figure*}

\begin{itemize}
  \item We propose an end-to-end pipeline of 3D point cloud instance segmentation with the input of projected spherical images.
  \item We generate the spherical images dataset from KITTI 3D Object Track dataset \cite{KITTI} with the corresponding instance labels to train our model. 
  \item We extend the Sort \cite{sort} to build the graph with the normalized distance of each object location center instead of only using IoU of the object detection bounding boxes to solve the path interaction problem. 
  \item We propose a stable and fast multi-object track structure with the extended Sort \cite{sort} and light-version 3D instance segmentation pipeline based on Mask RCNN \cite{MRCNN} and MobileNet structure \cite{mobilenet}.
\end{itemize}

\section{Related Work}
In this section, we focus on the recent approaches entirely for instance segmentation, multiple objects tracking and some data fusion methods which always used in popular 3D object detection models.
\subsection{Instance Segmentation}
Due to the significant development of bounding box object detection, many early instance segmentation approaches are based on two stages: (1) use box detector to generate a set of bounding boxes for the location of each possible object position; 
(2) predict the pixel-wise mask with the bounding boxes proposals and the feature maps which are sensitive to the instance location. 
For example, MNC \cite{MNC} followed the two stages to build a multiple-branch cascade model to generate accurate results from all the bounding-box proposals. However, this method is time-consuming which is not suitable for intelligent vehicles. For instance, MNC \cite{MNC} costs around 0.4s in the process of feature extraction.

Some recent approaches combine the segmentation methods and object detection system to achieve the instance segmentation with an extended fully convolutional network which can only be applied in the semantic segmentation task. 
For example, DeepMask \cite{deepmask} and FCIS \cite{fcis}, predicted the position-sensitive score maps from the extended Fully Convolutional Network (FCN). 
Those maps can locate the object and predict masks at the same time. However, they usually perform mistakes in mask prediction in the overlap of the adjacent instances \cite{deepmask} \cite{fcis}.

In this paper, we use Mask RCNN \cite{MRCNN}, different from SGPN \cite{SGPN}, as the pipeline which predicts the object bounding box and takes the corresponding pixel-wise mask parallelly from the existing shared branch. SGPN \cite{SGPN} performed 3D instance segmentation well in the indoor scene. 
But the problem still exists when it is applied in the outdoor large-scale scenes. A large quantity of points will cause memory inefficiency in \cite{SGPN}. 
To improve this, we introduce the spherical images generated based on the point clouds, and use this 2D projected data as the input to get the point-wise mask from the model outputs.

\subsection{Multiple Object Tracking}
Many popular multiple objects tracking methods consider the Multiple Object Tracking (MOT) problem as a data association problem and match the detection targets between multiple continuous frames.
For example, Sort used Kalman filter in the bounding box to generate the association matrix between two frames with the optimization of the assignment cost matrix using Hungarian algorithm \cite{sort}. 
However, the identity switch and tracked target loss are the problems in Sort \cite{sort} because its generation method of the association matrix will be influenced easily by the object intersections. 
Deepsort \cite{deepsort} used the recursive Kalman filtering and a more complex association matrix to solve the problem of a large number of identity switches. 
However, the deep association matrix can only be trained from the particular a large-scale object re-identification dataset which is not convenient for others to apply in different scenes. 

Moreover, some approaches for 3D tracking combine the RGB image information and 3D point clouds by data alignment to achieve the state estimation in 3D space. 
For example, 3D-CNN/PMBM used a deep learning structure to estimate the distance from the camera to object and combine the distance information to generate the association matrix \cite{PMBM}. 
In practice, these methods cannot provide stable results due to the inaccurate estimation in  intelligent vehicles. 
\subsection{Data Fusion in Deep Learning}
In deep learning, many researchers usually get the input from the multiple data sources to produce more consistent and robust features. 
MV3D used the bird view, the front view of the LiDAR and the corresponding RGB image to get enough feature descriptions for the 3D object detection \cite{MV3D}. 
Additionally, F-PointNet \cite{F-PointNet} combined the RGB image with the depth information to extract more stable feature representations.
PointSeg \cite{PointSeg} and SqueezeSeg \cite{SqueezeSeg} projected the 3D point clouds into the spherical coordinate system and performed semantic segmentation on the projected data. 
Moreover, VoxelNet \cite{VoxelNet} proposed a novel layer to learn useful features directly from the points in each voxel.

\section{Methodology}
This section firstly provides the details about the input of the proposed \textit{PointIT}, the main features of instance segmentation part. The rest of this section discusses the key points in the extend sort methods separately.
\begin{figure}[!h]
  \centering
    \includegraphics[width=0.4\textwidth]{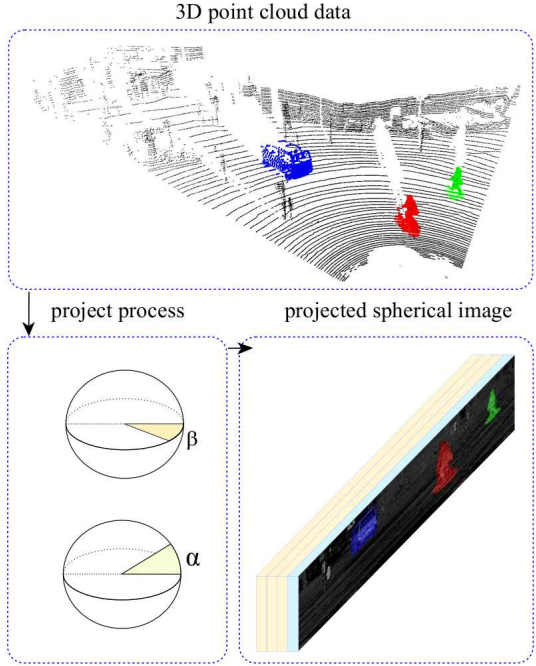}
  \caption{The process from 3D point clouds to sphearica image.
}
  \label{fig:spherical}
\end{figure}
\subsection{Network Input}
We find that PointSeg\cite{PointSeg} and SqueezeSeg   \cite{SqueezeSeg} perform semantic segmentation well in the spherical image and optimize the mask with the corresponding 3D information. 
We draw on their successes on utilizing the 3D point cloud data into the spherical image as the input of our \textit{PointIT}.
Different from the input of  \cite{PointSeg} and   \cite{SqueezeSeg}, we project the 3D LiDAR data into a spherical projected image with four channels, which are corresponding to the Cartesian coordinates information $(x, y, z)$ and the $reflectivity$.  
We transform the LiDAR data into the spherical image as follows:
\begin{equation}
\alpha = arcsin(\frac{z}{\sqrt{x^2+y^2+z^2}}) \ \ \  \bar{\alpha}=\lfloor\frac{\alpha}{\Delta\alpha}\rfloor,
\label{equ:azimuth}
\end{equation}
\begin{equation}
\beta =arcsin(\frac{y}{\sqrt{x^2+y^2}}) \ \ \ \bar{\beta}=\lfloor\frac{\beta}{\Delta\beta}\rfloor,
\label{equ:zenith}
\end{equation}
where $\alpha$ and $\beta$ are the azimuth and zenith angles, respectively see in Fig . \ref{fig:spherical} respectively. $\Delta \alpha$ and $\Delta \beta$ are the sizes of the spherical image which we want to generate. 
In our model, we set $\Delta \alpha = 64$ and $\Delta \beta = 512$. The $\bar{\alpha}$ and $\bar{\beta}$ are the position indexes on the projected image.
We will generate the array with the shape $64\times 512 \times 4$ as the input of our model. 
$64$ is set because of the 3D LiDAR data is come from the Velodyne HDL-64E LiDAR with 64 vertical channels,
while $512$ is set because we only project the front view area $(-45^\circ, 45^\circ)$ into the spherical image. 
After the transformation, we feed this image-type data into the instance segmentation part of the \textit{PointIT} to obtain the instance masks directly.

\subsection{Instance Segmentation} 
\begin{figure}[!h]
  \centering
    \includegraphics[width=0.48\textwidth]{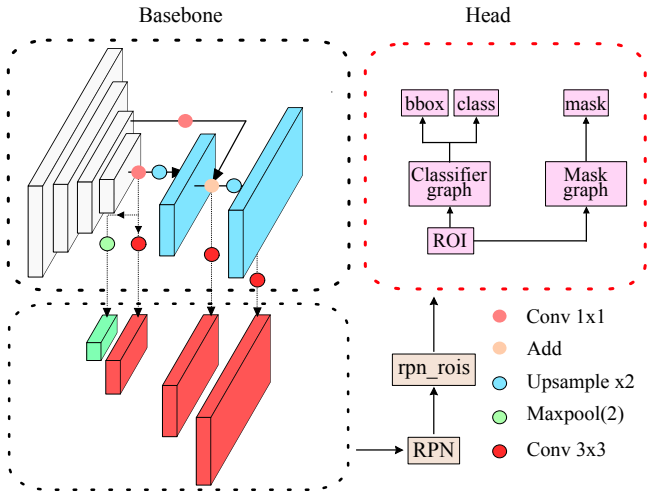}
  \caption{The structure of the instance segmentation model in the \textit{PointIT}. It uses the backbone (MobileNet) to extract the features and contains four different shapes of feature maps for RPN layer. 
}
  \label{fig:network}
\end{figure}

In order to achieve a satisfactory efficiency, we build a fast instance segmentation model followed the MobileNet \cite{mobilenet} and Mask RCNN \cite{MRCNN}.
\begin{table}[h]
\center
\caption{The parameters of the encoder.}

\renewcommand\arraystretch{1.5}
   \renewcommand\tabcolsep{7.4pt}
\begin{tabular}{@{}ccllll@{}}
\toprule
{Input} & {Operation}             & {c}  & {s} \\ \midrule
H$\times$W$\times$4          & conv2d                         & 32         & 2          \\ \midrule
H/2$\times$W/2$\times$32     & depthwise\_separable\_block & 64          & 1          \\ \midrule
H/2$\times$W/2$\times$64       & depthwise\_separable\_block & 128         & 2         \\ \midrule
H/4$\times$W/4$\times$128       & depthwise\_separable\_block & 128         & 1         \\ \midrule
H/4$\times$W/4$\times$128     & depthwise\_separable\_block & 256         & 2          \\ \midrule
H/8$\times$W/8$\times$256     & depthwise\_separable\_block & 256         & 1          \\ \midrule
H/8$\times$W/8$\times$256   & depthwise\_separebla\_block & 512           & 2          \\ \midrule
H/16$\times$W/16$\times$512   & depthwise\_separable\_block & 512           & 1          \\ \midrule
H/16$\times$W/16$\times$512   & depthwise\_separable\_block & 512           & 1          \\ \midrule
H/16$\times$W/16$\times$512   & depthwise\_separable\_block & 512           & 1          \\ \midrule
H/16$\times$W/16$\times$512   & depthwise\_separable\_block & 512           & 1          \\ \midrule
H/16$\times$W/16$\times$512   & depthwise\_separable\_block & 512           & 1          \\ \midrule
\end{tabular}
\label{table:para}
\end{table}
MobileNet\cite{mobilenet} proposed a novel layer function called the depthwise separable convolution block, which is constructed with the depthwise convolutional layer and the pointwise convolutional layer. 
Using this function, the model can keep the balance between the performance of accuracy and efficiency.
In this paper, we take this function as the unit of the feature extractor of our model and more details can be read in \cite{mobilenet}.
The parameters of the encoder are shown in Table. \ref{table:para}. The \textit{c} is the output channels of the operation, and the \textit{s} is the stride size. 
The shape of the input will be downsampled four times. The end feature will have the dimension $16/H\times16/W \times512$. 
We do not add more downsampling process in the encoder because the Feature Pyramid Networks (FPN) features \cite{fpn} are generated from the end feature maps in the encoder with the max-pooling layer, which can be seen as the green box in Fig. \ref{fig:network}. 
Also we set the total $stride=16$ to save more information and avoid the feature maps generating $height=2$.

The structure of the instance segmentation module can be seen in Fig. \ref{fig:network}. It contains four downsampling operations in the encoder. 
Also RPN layers will generate the ROIs(region of interest area) based on the four different scales of the feature maps. 
Then the network head will build the classifier graph and mask graph parallelly. The classifier graph is constructed with one convolutional layer with a $7\times 7$ kernel size and two $1\times 1$ convolutional layers to generate the feature maps for the object classification and object bounding box location.
The mask graph is built with the convolutional and deconvolutional layers to create the mask map with the shape of $28\times 28$.

\subsection{Extended Sort}
To extend it into the 3D tracking framework, we adopt the space information which contains the $[x, y, z]^{\top}$ in the projected data with the recursive Kalman filtering.
And we build a more stable association matrix between the frames than the original function in   \cite{sort}. We will describe the details in the following.
\subsubsection{State Estimation}
To estimate the object motion, we build the motion model to predict the location of the target in the next frame. The state of each target is generated as:
\begin{align}
&X_{Object} = (x_p,y_p,s,r,\bar{x}, \bar{y}, \bar{s})^{\top} \label{equ:ob2d}\\
&X_{Center} =(x_w, y_w, z_w, \bar{V_x}, \bar{V_y}, \bar{V_z}, \bar{A_x}, \bar{A_y})^{\top} \label{equ:ob3d}
\end{align}
The Eq. \ref{equ:ob2d} contains the center of a bounding box position $[x_p, y_p]^{\top}$, its scale $s$, and the aspect ratio $r$. We use Kalman filter with the constant velocity and take the $[x_p,y_p,s,r]^{\top}$ as the observations of the object state. The Eq. \ref{equ:ob3d} has the centre location $[x_w, y_w, z_w]^{\top}$ of the object in the 3D space, the velocity $[\bar{V_x}, \bar{V_y}, \bar{V_z}]^{top}$ and acceleration $[\bar{A_x}, \bar{A_y}]^{\top}$. The Kalman filter is taken with the constant velocity in $z$ and has the acceleration in $x$ and $y$ to smooth the estimation. And the observation state is the $[x_w, y_w, z_w]^{\top}$. 
\subsubsection{Data Assignment Problem}
To assign the tracked objects in each frame, we estimate the object bounding box and the object location in the world space based on the current frame.
We build the cost matrix by the weighted IoU between and the weighted distance. The  weighted IoU is calculated by all current detected bounding boxes and the new predicted bounding boxes. The weighted distance is generated from all detected target locations and the estimated location of all targets. All functions are shown in Eqs. \ref{equ:graph}, \ref{equ:IoU}, \ref{equ:dis}. Then we use Hungarian algorithm to solve this assignment problem to get the minimum cost. The graph functions are shown as follows:
\begin{align}
&Graph(i,j) = \alpha \cdot I(i,j) + \beta \cdot D(i,j) \label{equ:graph}\\
&	I(i,j) = \frac{Box_i \bigcap Box_j}{Box_i \bigcup Box_j} \label{equ:IoU}\\	
& D(i,j) = \exp[-dis(i,j)] \\
& dis		  =\sqrt[2]{||P_i -P_j||} \ \ \ P=[x,y,z]^{\top} \label{equ:dis}
\end{align}
In the Eq. \ref{equ:graph}, $\alpha + \beta =1$ and they are used to weight the balance of two matrixes. We set the $\alpha$ as 0.5 to assume that they have the same weight. $I$ means the IoU matrix. $D$ represents the distance matrix. The values of $I$ and $D$ are $\in[0,1]$. $i$ represents the detection target and $j$ represents the prediction target with the state estimation. The details of the process are described in the Algorithm.\ref{alg:exsort}.

\begin{algorithm}
\caption{Extend Sort process.}
	\begin{algorithmic}[l]
		\STATE \textbf{Input of frame $t$ : } {Detection box $B_t=\left(B_1,B_2,...,B_i\right)$};
		Mask $M_t =\left(M_1, M_2,...,M_i\right)$; The estimated target $E_t=\left(E_1,E_2,...,E_j\right)$.
		\STATE \textbf{Output of frame $t$ :} matched indices $M$, unmatched indices $UM$ and the estimation state of frame $t+1$.
		\STATE \textbf{1:} Compute center location $L_t$ with the corresponding Mask: $L_t = \left(L_1,L_2,...,L_i\right)$
		\STATE \textbf{2:} Build the cost matrix $Cost_t$ with the Eq. \ref{equ:graph}
		\STATE \textbf{3:} Initialize the set of matched indices $M\gets 0 $ and the set of unmatched indices $UM\gets D_t$
		\STATE \textbf{4:}  $MID_{t}, UID_{t} \gets$ $Minimize(Cost_{t})$
		\STATE \textbf{5:}	 Remove $ID$ in $UM$ for $ID$ in $MID_{t+1}$
		\STATE \textbf{6:} Add $ID$ in $M$ for $ID$ in $UID_{t+1}$
	\end{algorithmic}
	
\label{alg:exsort}
\end{algorithm}
In Algorithm.\ref{alg:exsort}, we describe the whole process of the proposed extend sort and assume the task starts with the frame $t$. Then in frame $t+1$, the matched and unmatched indices
in step 3 will not be initialized again and just take the tracking results from the previous frame.




	 


	

	
	
	
	
	



\begin{table*}[t]
\center
\caption{Performance of the propsed approach on generated sequence}
\scalebox{1.3}{
\begin{tabular}{@{}clcccccccc@{}}
\toprule
\textbf{Method}       & \textbf{Type} & \textbf{MOTA}$\uparrow$        & \textbf{MOTP}$\uparrow$         & \textbf{MT}$\uparrow$           & \textbf{ML}$\downarrow$           & \textbf{ID\_sw}$\downarrow$      & \textbf{FM}$\downarrow$          & \textbf{FP}     $\downarrow$     & \textbf{FN}$\downarrow$          \\ \midrule
\multicolumn{1}{l}{baseline (sort)}             & Online        & 0.451                 & 0.80                 & 0.137                 & 0.379                 & 5                   & 50                   & 836                  & 1945                 

\\

\multicolumn{1}{l}{our proposed} & Online        & \textbf{0.457} &0.80 &\textbf{0.155} & \multicolumn{1}{l}{\textbf{0.327}} & \textbf{2} & 56 & 895 & \textbf{1895} \\ \bottomrule
\end{tabular}}
\label{tabel:sort}

\end{table*}

\section{Experiments}
All the experiments are conducted on server equipped with one NVIDIA GeForce GTX GPU with the CUDA 9 and CUDNN v7. During the training model of instance segmentation, we set the learning rate as 0.0001. We train the network for 2000 ephos, which has 500 steps in each epho.
\subsection{Datasets and Evaluation Results}
We first train the model of the instance segmentation task on the generated dataset which is transformed from the KITTI Track Object dataset \cite{KITTI}. We split the generated dataset into two forms. One part (instance dataset) which from sequence '0000' to sequence '0018', is used to train the instance segmentation model. The other sequences '0019',  '0020' are used for the testing of tracking. 
Also, there are 5000 frames in the generated instance dataset and we separate it into on training set with 4500 frames and one evaluation set with 500 frames.

\subsubsection{Evaluation of the Instance Segmentation on Generated Dataset}
\begin{figure}[!h]
  \centering
    \includegraphics[width=0.45\textwidth]{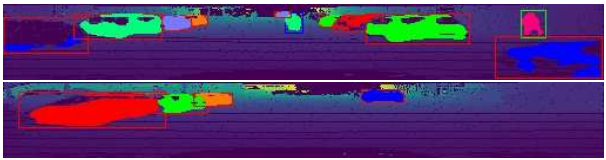}
  \caption{The results of the instance segmentation on the generated data without tracking id. The red bounding box represents the car, the blue bounding box represents the cyclist, and the pedestrian is shown by the green bounding box.
}
  \label{fig:ins_image}
\end{figure}
Table \ref{table:runtime} shows the evaluation results of the instance segmentation part of our model. It shows a good performance in the projected point data and the AP (average precision) achieves  $0.617$ among the evaluation set when the IoU threshold is 0.5. We evaluate the testing dataset with the threshold 0.7, and the AP is only $0.237$. The reason for this gap between the two thresholds is that the input shape is quite small. When the segmented objects are far away from the camera, the slight change in the detected box will make a significant influence on the calculation of IoU of the detected bounding box and the ground truth. The simple results of the instance segmentation have been shown in Fig. \ref{fig:ins_image}.

The comparison of runtime and average precision has been shown in Table \ref{table:runtime}, 
a light-version structure (Mobilenet) with the instance segmentation baseline can achieve similar performance as the ResNet backbone \cite{resnet} in the projected spherical image. However, the light-version structure can save more memory and computation time in the process.

\begin{table}[t]
	\caption{The performance comparsion in different backbone in runtime and average precision.}

	\scalebox{1.3}{
	\begin{tabular}{cccc}
		\toprule
backbone   & runtime(s) & AP\_0.5 & AP\_0.7 \\\midrule
Resnet\_50 & 0.091           & 0.66        &0.251         \\ 
Mobilenet  & 0.061       & 0.617   & 0.237   \\ \midrule	
	\end{tabular}}
			\label{table:runtime}
\end{table}

\subsubsection{Evaluation of PointIT}
We evaluate the performance of our tracking model with the multi-target scores. The carried evaluation metrics include:
\begin{itemize}
 \item MOTA\cite{mota}: multi-object tracking accuracy.
 \item MOTP\cite{mota}: multi-object tracking precision.
 \item MT: number of mostly tracked trajectoris when the life span of tracked targed is larger than 80\%.
 \item ML: number of mostly tracked trajectoris when the life span of tracked targed is smaller than 20\%.
 \item ID\_sw: number of times when an ID changes into a different tracked object.
 \item FM: number of the times when a track is lost due to the missing detection.
\end{itemize}
In the evaluation measurement, $(\uparrow)$ repesents the higher score will perform better in the task and the $(\downarrow)$ denotes the lower score will achieve better performance.

\begin{figure*}[!h]
  \centering
    \includegraphics[width=1\textwidth]{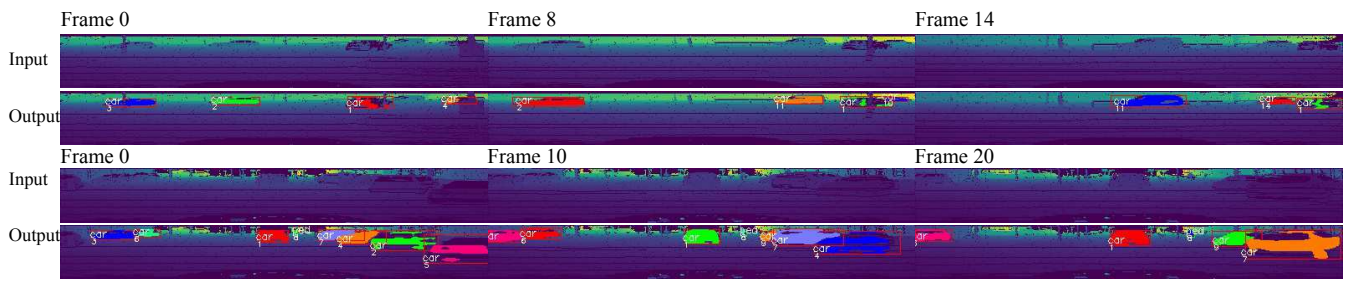}
  \caption{The results with the track id are shown among the one sequence in each row. The red boxes represent cars and the green box represents the pedestrians. The colors of masks are set only for  visualization.
}
  \label{fig:result}
\end{figure*}

Table \ref{tabel:sort} only shows the evaluation results on the car due to the limitation of the generated testing dataset which mostly contains the object car in the whole sequence. 
From Table \ref{tabel:sort}, we only compare the results with the Sort \cite{sort}. 
With the additional information from 3D space, the performance of our method shows a good improvement in the \textbf{MT}, \textbf{ML} and \textbf{FN} and \textbf{ID\_sw} in the testing dataset. 
In the spherical image, the indexes are calculated with the resolution of height and width. The problem of the occlusion has eased in this projection. Because of this, the score of ID sw is low in the whole testing sequence.
We do not compare with other state-of-the-art methods, such as DeepSort  \cite{deepsort} which we can not generate a relevant dataset to train the deep association matrix, 
and our method is proposed to find an efficient way to combine the space information into the 3D object tracking task. 
The track results are shown in Fig. \ref{fig:result}, each of the sequences is chosen the three frames to show the performance of the track framework and do not show the 3D point data visualization in the paper.
\section{Discussion}
Our \textit{PointIT} achieves impressive results which show that tracking the object in 3D space is more accurate. 
We propose a simple way to introduce the spatial information into the existing popular track frameworks which are based on the bounding box detectors directly, like Sort \cite{sort}. 
With the additional spatial information, not only the tracking methods, but also other frameworks can cooperate well with our proposed 3D instance segmentation framework. 


From the Table \ref{table:runtime}, we can find that the AP$_{0.5}$ of Resnet only improved 4\% than the Mobilenet, which is not expected by us. We think that there are some different components between the standard 2D image and the projected spherical data, such as the domain distribution. We assume that the deeper structure of the network would influence its feature learning and the imbalance data distribution will cause an upper bound limitation of the performance.

We filter the LiDAR from 360 degrees into the 90 degrees because the front view of one vehicle is the most important part which the vehicles need to care. Our projected data does not contain the distance which is included in  \cite{PointSeg} and  \cite{SqueezeSeg} because we find that there is not a huge difference between two different implementations. Also in the ideal situation, all the indexes in the spherical projected data should have different corresponding points with $x,y,z$ and $reflectivity$. However, the noise is in the 3D point clouds and the LiDAR will lose some points due to the intensity or material of the object surface. Both of the problems cause the loss of points in the projected spherical and the irregular mask in the predict results which is shown in Fig. \ref{fig:ins_image}.  
\section{Conclusion}
In this paper, we proposed a fast online tracking framework based on 3D point clouds. 
We constructed a light-weight structure to achieve the 3D point cloud instance segmentation on the road objects. 
In addition, we trained our instance prediction model on generated datasets which were generated from the KITTI Object Track dataset. 
We showed that with the help of the accurate space information, e.g. the centre point of the 3D instance object, the tracking performance could be improved hugely. 
The proposed \textit{PointIT} greatly minimized the trade-off between the performance and the efficiency in the 3D object tracking challenge. 

\bibliographystyle{IEEEtran}
\bibliography{root}

\begin{thebibliography}{10}
\providecommand{\url}[1]{#1}
\csname url@rmstyle\endcsname
\providecommand{\newblock}{\relax}
\providecommand{\bibinfo}[2]{#2}
\providecommand\BIBentrySTDinterwordspacing{\spaceskip=0pt\relax}
\providecommand\BIBentryALTinterwordstretchfactor{4}
\providecommand\BIBentryALTinterwordspacing{\spaceskip=\fontdimen2\font plus
\BIBentryALTinterwordstretchfactor\fontdimen3\font minus
  \fontdimen4\font\relax}
\providecommand\BIBforeignlanguage[2]{{%
\expandafter\ifx\csname l@#1\endcsname\relax
\typeout{** WARNING: IEEEtran.bst: No hyphenation pattern has been}%
\typeout{** loaded for the language `#1'. Using the pattern for}%
\typeout{** the default language instead.}%
\else
\language=\csname l@#1\endcsname
\fi
#2}}

\bibitem{sort}
\BIBentryALTinterwordspacing
A.~Bewley, Z.~Ge, L.~Ott, F.~Ramos, and B.~Upcroft, ``Simple online and
  realtime tracking,'' \emph{CoRR}, vol. abs/1602.00763, 2016. [Online].
  Available: \url{http://arxiv.org/abs/1602.00763}
\BIBentrySTDinterwordspacing

\bibitem{deepsort}
\BIBentryALTinterwordspacing
N.~Wojke, A.~Bewley, and D.~Paulus, ``Simple online and realtime tracking with
  a deep association metric,'' \emph{CoRR}, vol. abs/1703.07402, 2017.
  [Online]. Available: \url{http://arxiv.org/abs/1703.07402}
\BIBentrySTDinterwordspacing

\bibitem{vojir2014robust}
T.~Vojir, J.~Noskova, and J.~Matas, ``Robust scale-adaptive mean-shift for
  tracking,'' \emph{Pattern Recognition Letters}, vol.~49, pp. 250--258, 2014.

\bibitem{Complex-YOLO}
\BIBentryALTinterwordspacing
M.~Simon, S.~Milz, K.~Amende, and H.~Gross, ``Complex-yolo: Real-time 3d object
  detection on point clouds,'' \emph{CoRR}, vol. abs/1803.06199, 2018.
  [Online]. Available: \url{http://arxiv.org/abs/1803.06199}
\BIBentrySTDinterwordspacing

\bibitem{MV3D}
X.~Chen, H.~Ma, J.~Wan, B.~Li, and T.~Xia, ``Multi-view 3d object detection
  network for autonomous driving,'' in \emph{IEEE CVPR}, vol.~1, no.~2, 2017,
  p.~3.

\bibitem{VoxelNet}
Y.~Zhou and O.~Tuzel, ``Voxelnet: End-to-end learning for point cloud based 3d
  object detection,'' \emph{arXiv preprint arXiv:1711.06396}, 2017.

\bibitem{PIXOR}
B.~Yang, W.~Luo, and R.~Urtasun, ``Pixor: Real-time 3d object detection from
  point clouds,'' in \emph{Proceedings of the IEEE Conference on Computer
  Vision and Pattern Recognition}, 2018, pp. 7652--7660.

\bibitem{SqueezeSeg}
\BIBentryALTinterwordspacing
B.~Wu, A.~Wan, X.~Yue, and K.~Keutzer, ``Squeezeseg: Convolutional neural nets
  with recurrent {CRF} for real-time road-object segmentation from 3d lidar
  point cloud,'' \emph{CoRR}, vol. abs/1710.07368, 2017. [Online]. Available:
  \url{http://arxiv.org/abs/1710.07368}
\BIBentrySTDinterwordspacing

\bibitem{PointSeg}
\BIBentryALTinterwordspacing
Y.~Wang, T.~Shi, P.~Yun, L.~Tai, and M.~Liu, ``Pointseg: Real-time semantic
  segmentation based on 3d lidar point cloud,'' \emph{CoRR}, vol.
  abs/1807.06288, 2018. [Online]. Available:
  \url{http://arxiv.org/abs/1807.06288}
\BIBentrySTDinterwordspacing

\bibitem{pointnet}
C.~R. Qi, H.~Su, K.~Mo, and L.~J. Guibas, ``Pointnet: Deep learning on point
  sets for 3d classification and segmentation,'' \emph{Proc. Computer Vision
  and Pattern Recognition (CVPR), IEEE}, vol.~1, no.~2, p.~4, 2017.

\bibitem{KITTI}
A.~Geiger, P.~Lenz, and R.~Urtasun, ``Are we ready for autonomous driving? the
  kitti vision benchmark suite,'' in \emph{Computer Vision and Pattern
  Recognition (CVPR), 2012 IEEE Conference on}.\hskip 1em plus 0.5em minus
  0.4em\relax IEEE, 2012, pp. 3354--3361.

\bibitem{MRCNN}
\BIBentryALTinterwordspacing
K.~He, G.~Gkioxari, P.~Doll{\'{a}}r, and R.~B. Girshick, ``Mask {R-CNN},''
  \emph{CoRR}, vol. abs/1703.06870, 2017. [Online]. Available:
  \url{http://arxiv.org/abs/1703.06870}
\BIBentrySTDinterwordspacing

\bibitem{mobilenet}
\BIBentryALTinterwordspacing
A.~G. Howard, M.~Zhu, B.~Chen, D.~Kalenichenko, W.~Wang, T.~Weyand,
  M.~Andreetto, and H.~Adam, ``Mobilenets: Efficient convolutional neural
  networks for mobile vision applications,'' \emph{CoRR}, vol. abs/1704.04861,
  2017. [Online]. Available: \url{http://arxiv.org/abs/1704.04861}
\BIBentrySTDinterwordspacing

\bibitem{MNC}
J.~Dai, K.~He, and J.~Sun, ``Instance-aware semantic segmentation via
  multi-task network cascades,'' in \emph{Proceedings of the IEEE Conference on
  Computer Vision and Pattern Recognition}, 2016, pp. 3150--3158.

\bibitem{deepmask}
\BIBentryALTinterwordspacing
P.~H.~O. Pinheiro, R.~Collobert, and P.~Doll{\'{a}}r, ``Learning to segment
  object candidates,'' \emph{CoRR}, vol. abs/1506.06204, 2015. [Online].
  Available: \url{http://arxiv.org/abs/1506.06204}
\BIBentrySTDinterwordspacing

\bibitem{fcis}
\BIBentryALTinterwordspacing
Y.~Li, H.~Qi, J.~Dai, X.~Ji, and Y.~Wei, ``Fully convolutional instance-aware
  semantic segmentation,'' \emph{CoRR}, vol. abs/1611.07709, 2016. [Online].
  Available: \url{http://arxiv.org/abs/1611.07709}
\BIBentrySTDinterwordspacing

\bibitem{SGPN}
\BIBentryALTinterwordspacing
W.~Wang, R.~Yu, Q.~Huang, and U.~Neumann, ``{SGPN:} similarity group proposal
  network for 3d point cloud instance segmentation,'' \emph{CoRR}, vol.
  abs/1711.08588, 2017. [Online]. Available:
  \url{http://arxiv.org/abs/1711.08588}
\BIBentrySTDinterwordspacing

\bibitem{PMBM}
\BIBentryALTinterwordspacing
S.~Scheidegger, J.~Benjaminsson, E.~Rosenberg, A.~Krishnan, and
  K.~Granstr{\"{o}}m, ``Mono-camera 3d multi-object tracking using deep
  learning detections and {PMBM} filtering,'' \emph{CoRR}, vol. abs/1802.09975,
  2018. [Online]. Available: \url{http://arxiv.org/abs/1802.09975}
\BIBentrySTDinterwordspacing

\bibitem{F-PointNet}
C.~R. Qi, W.~Liu, C.~Wu, H.~Su, and L.~J. Guibas, ``Frustum pointnets for 3d
  object detection from rgb-d data,'' \emph{arXiv preprint arXiv:1711.08488},
  2017.

\bibitem{fpn}
\BIBentryALTinterwordspacing
T.~Lin, P.~Doll{\'{a}}r, R.~B. Girshick, K.~He, B.~Hariharan, and S.~J.
  Belongie, ``Feature pyramid networks for object detection,'' \emph{CoRR},
  vol. abs/1612.03144, 2016. [Online]. Available:
  \url{http://arxiv.org/abs/1612.03144}
\BIBentrySTDinterwordspacing

\bibitem{resnet}
K.~He, X.~Zhang, S.~Ren, and J.~Sun, ``Deep residual learning for image
  recognition,'' in \emph{Proceedings of the IEEE conference on computer vision
  and pattern recognition}, 2016, pp. 770--778.

\bibitem{mota}
K.~Bernardin and R.~Stiefelhagen, ``Evaluating multiple object tracking
  performance: the clear mot metrics,'' \emph{Journal on Image and Video
  Processing}, vol. 2008, p.~1, 2008.

\end{thebibliography}

\end{document}